\documentclass[letterpaper, 10 pt, conference]{ieeeconf}  

\usepackage{xcolor}
\usepackage{graphicx}
\usepackage{glossaries}
\usepackage{amsmath}
\usepackage{amssymb}
\usepackage{multirow, makecell}
\usepackage{siunitx}
\usepackage{url}
\usepackage{bm}
\let\labelindent\relax
\usepackage{enumitem}

\IEEEoverridecommandlockouts                              

\title{\LARGE \bf
Learning Quasi-Static 3D Models of Markerless Deformable Linear Objects for Bimanual Robotic Manipulation\\
}

\author{Piotr Kicki, Michał Bidziński and Krzysztof Walas
\thanks{This work was supported by the European Union's Horizon 2020 Research and Innovation Programme under grant agreement No 870133, REMODEL.}
\thanks{All authors are with the Institute of Robotics and Machine Intelligence,
        Poznan University of Technology, Poznan, Poland
        {\tt\small \{name.surname\}@put.poznan.pl}}%
}

\begin{document}

\newcommand{\state}{\bm{s}}
\newcommand{\statepoc}{\state^{PoC}}
\newcommand{\statecp}{\state^{BsCP}}
\newcommand{\eepose}{\bm{p}}
\newcommand{\trans}{\bm{t}}
\newcommand{\rot}{\bm{R}}
\newcommand{\curv}{c}
\newcommand{\cp}{\alpha}
\newcommand{\curvpar}{\zeta}
\newcommand{\idx}{n}
\newcommand{\reals}{\mathbb{R}}
\newcommand{\numPts}{n_s}
\newcommand{\numAct}{n_a}
\newcommand{\action}{\bm{a}}
\newcommand{\jac}{J}

\newcommand{\pare}[1]{\left({#1}\right)}

\newacronym{poc}{PoC}{Points on Curve}
\newacronym{cp}{BsCP}{B-spline Control Points}
\newacronym[plural=DLOs, firstplural=Deformable Linear Objects (DLOs)]{dlo}{DLO}{Deformable Linear Object}
\newacronym{referenceFrame}{RF}{reference frame}
\newacronym[plural=NNs, firstplural=neural networks (NNs)]{nn}{NN}{neural network}
\newacronym[plural=EEs, firstplural=end-effectors (EEs)]{ee}{EE}{end-effector}

\newacronym{mlp}{MLP}{Multilayer Perceptron}

\newacronym{cem}{CEM}{Cross Entropy Method}
\newacronym{fem}{FEM}{Finite Element Method}

\newacronym{pbd}{PBD}{Position-Based Dynamics}

\newacronym{inbilstm}{IN-biLSTM}{interaction network with bidirectional LSTM}

\maketitle
\thispagestyle{empty}
\pagestyle{empty}

\begin{abstract}
The robotic manipulation of \glspl{dlo} is a vital and challenging task that is important in many practical applications. Classical model-based approaches to this problem require an accurate model to capture how robot motions affect the deformation of the \gls{dlo}. Nowadays, data-driven models offer the best tradeoff between quality and computation time. This paper analyzes several learning-based 3D models of the \gls{dlo} and proposes a new one based on the Transformer architecture that achieves superior accuracy, even on the \glspl{dlo} of different lengths, thanks to the proposed scaling method. Moreover, we introduce a data augmentation technique, which improves the prediction performance of almost all considered \gls{dlo} data-driven models. Thanks to this technique, even a simple \gls{mlp} achieves close to state-of-the-art performance while being significantly faster to evaluate. 
In the experiments, we compare the performance of the learning-based 3D models of the \gls{dlo} on several challenging datasets quantitatively and demonstrate their applicability in the task of shaping a \gls{dlo}.

\end{abstract}

\section{Introduction}
People encounter and skillfully manipulate \glspl{dlo}, such as cables, ropes, threads, strings, and hoses. It would be beneficial to give robots similar skills to enable them to perform cable routing~\cite{wlatersson2022routing, tomizuka2022routing}, knot tying~\cite{knots}, rope untangling~\cite{priya2021untangling}, belt-drive unit assembly~\cite{tomizuka2021optimization}, wiring harness assembly in the automotive sector~\cite{zhu2020contacts, wiring_harness_assembly}, threading the lace through the narrow hole~\cite{berenson2015tightholes}, or surgical suturing~\cite{surgical_sutures}. A typical approach to manipulating \glspl{dlo} is to use their models to plan the motion and control commands necessary to rearrange it to the desired state.
In recent years, there were many attempts to develop \gls{dlo} models, such as FEM-based ones~\cite{FEM, apetit_siciliano, coros2018interactive, coros2021dynamic}, using Cosserat rod theory~\cite{cosserat_rod, lang2011, kavan2016cosserat}, Kirchoff elastic rod~\cite{bretl2014, geometricallyexact2022} or dynamic B-splines~\cite{gianluca_ICPS, khalifa2022}.
However, these models make strong assumptions about the properties of the objects or require significant amounts of computations to evaluate, which limits their applicability in real-time manipulation or makes their relevance to the real \glspl{dlo} questionable.
A response to these problems was the trials to develop neural network-based models~\cite{stoyanov2022, mitrano2021trust, mingrui2023tro, wang2022dlomodelgnn, bohg2020prediction, wenbo2021linearlatentdynamics}. However, most of these methods are limited to manipulating DLOs on the plane~\cite{wang2022dlomodelgnn, bohg2020prediction, wenbo2021linearlatentdynamics}. In contrast, the ones that work in 3D typically utilize purposefully placed markers attached to the \gls{dlo} to get its exact state~\cite{stoyanov2022, mingrui2023tro}, which is an unrealistic assumption for the \gls{dlo} manipulation in the real world. Moreover, the inference times of the architectures proposed in~\cite{stoyanov2022} and~\cite{wang2022dlomodelgnn} are prohibitively long for real-time manipulation planning.

\begin{figure}[t]
    \centering
    \includegraphics[width=0.98\linewidth]{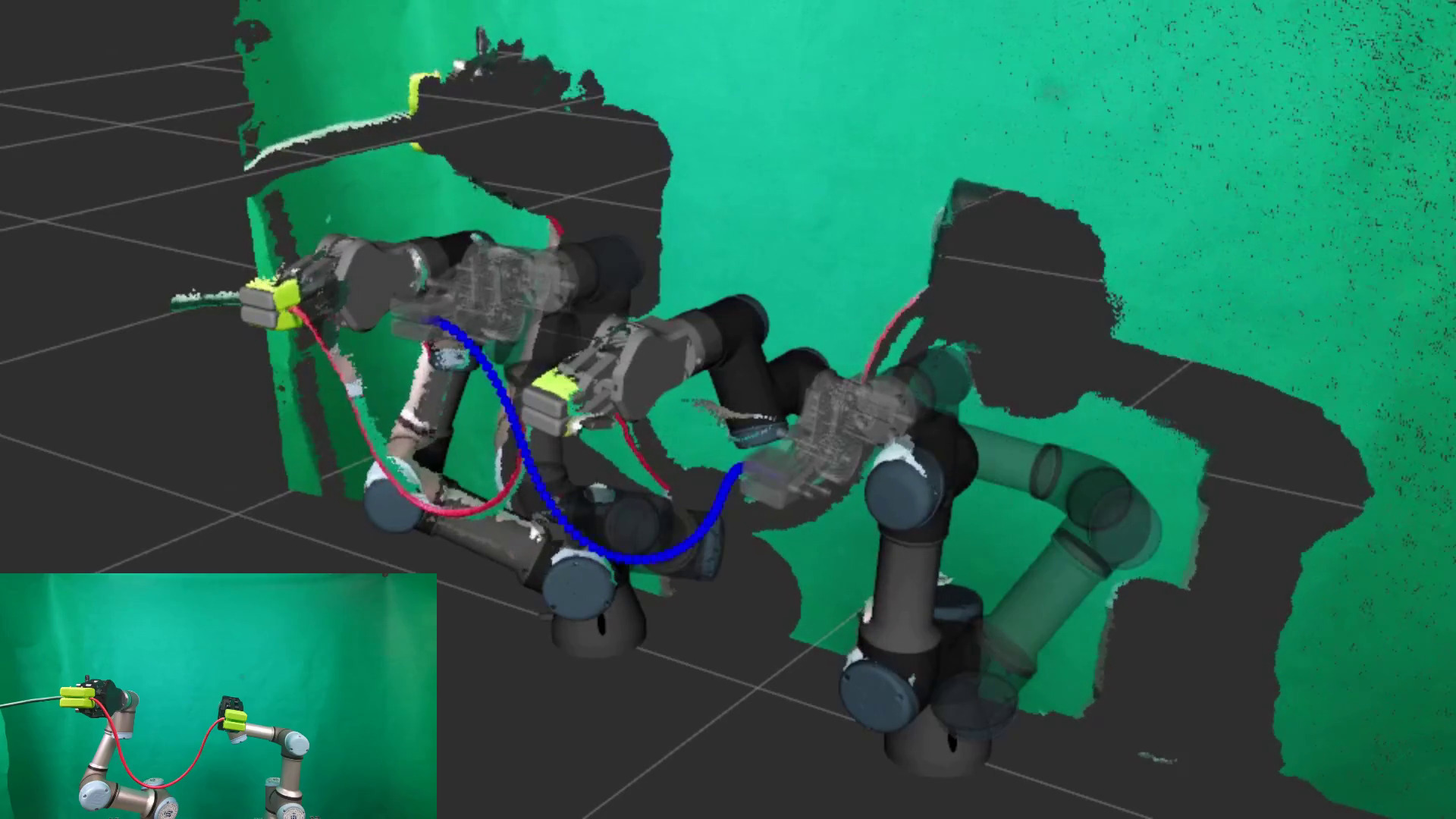}
    \caption{The proposed \gls{dlo} model's prediction of the markerless \gls{dlo} shape after a virtual move of the two UR3 robotic arms.}
    \label{fig:first}
\end{figure}

In this paper, we want to overcome the limitations mentioned above and present a neural network-based approach for a fast quasi-static model of the real \gls{dlo} being manipulated by a pair of robots in 3D space without the use of any markers (see~Figure~\ref{fig:first}).
To do so, the \gls{dlo} shape is being tracked with an RGBD camera with the improved version of our approach proposed in~\cite{dloftbs}. Based on tracking output, we train a machine learning model to be able to predict the state of a \gls{dlo} after the robot's move, given an actual \gls{dlo} state and an initial and final pose of the \glspl{ee} holding the \gls{dlo}. Our four contributions that pertain to achieving this goal are the following.

First, we propose to model the \gls{dlo} using a neural network with the Transformer architecture~\cite{transformer}. The attention mechanism shows superb prediction accuracy w.r.t. state-of-the-art learning-based \gls{dlo} models. In the literature, three main neural network architectures are reported: \gls{mlp}, interaction network~\cite{interaction2016networks}, and radial basis function network~\cite{saha1993rbfn}. Our approach with the Transformer is new in this field.
 
Second, we introduce a simple yet effective data augmentation procedure that significantly improves the accuracy of almost all considered architectures and enables the \gls{mlp} to achieve accuracy similar to much more complex models like \gls{inbilstm}~\cite{stoyanov2022} or Transformer while being substantially more computationally efficient.

Third, we analyzed the impact of different data representations on the prediction accuracy of neural network-based \gls{dlo} models.

Finally, we show that using the \gls{mlp} with the proposed data augmentation and \gls{cem} for motion prediction, one can plan the coordinated movement of two arms, which leads to a desired movement of the \gls{dlo}.

Furthermore, it is essential to stress the fact that the knowledge gained for one \gls{dlo} is transferable for the same \gls{dlo} type but different lengths thanks to the proposed data scaling and to \glspl{dlo} of different physical parameters. Furthermore, we evaluated the possibility of retraining a model pre-trained on a different \gls{dlo} setup. To the best of the authors' knowledge, this is the first analysis of the transferability of the neural \gls{dlo} models.
To facilitate the research on \gls{dlo} modeling, we make our code and datasets publicly available at \url{https://github.com/PPI-PUT/neural_dlo_model}.



\section{Related work}
\subsection{Physics-based DLO models}
Early models of deformable linear objects were designed based on the analysis of the physics of the phenomenons that govern the motion of the \gls{dlo}.
In the literature, there are several approaches to physics-based \gls{dlo} modeling, like (i) \gls{fem}~\cite{yoshida2015ringshapefem, coros2018interactive}, (ii) continuous elastic rod models~\cite{geometricallyexact2008, bretl2014}, (iii) multi-body models~\cite{umetani2014pbd, liu2023pbd}, or (iv) jacobian-based models~\cite{berenson2013jacobian, berenson2015tightholes}, each of which with its pros and cons~\cite{lv2020survey, kragic2021survey}.


Starting from FEM, the main advantage of mesh-based models is their accuracy and ability to simulate large deformations accurately. In~\cite{yoshida2015ringshapefem}, \gls{fem}-based model was used for simulation and planning of the ring-shape object deformation, while in~\cite{coros2018interactive} \gls{fem}-based model with sensitivity analysis was a key component of bi-manual robotic shape control of the deformable objects. The high accuracy of these methods is at the cost of long simulation times caused by huge amounts of computations.
To address that, authors of~\cite{koessler2021icra} proposed a method for shape control based on \gls{fem}, which does not require real-time simulation of the
deformable object in the control algorithm. To achieve real-time performance, they simplified the model of the manipulated deformable object to a 1D case, i.e., a chain of nodes.

A chain of nodes is also at the core of \gls{pbd}~\cite{umetani2014pbd, kavan2016cosserat}, constrained rigid bodies-based~\cite{servin2008rigidbodysequence}, and mass-spring~\cite{lv2017massspring} approaches to \gls{dlo} modeling. However, these approaches produce visually plausible animations instead of physically accurate simulations. To address this issue, in~\cite{liu2023pbd}, a compliant version of \gls{pbd} was used to model real rope-like objects. Nevertheless, these approaches are characterized by limited accuracy, especially for more stiff \glspl{dlo}, like cables.

An alternative approach is to represent the \gls{dlo} as a curve, which better resembles the nature of the \gls{dlo} and, at the same time, is less complex than volumetric mesh models. 
The key concept in \gls{dlo} modeling using elastic rods is that stable configurations of the \gls{dlo} correspond with minimal-energy curves that represent them~\cite{moll2006minimalenergy}.
This allows for more efficient \gls{dlo} path planning~\cite{moll2006minimalenergy}, perception~\cite{javdani2011} and manipulation~\cite{bretl2014}. The most popular approaches exploit the models based on Kirchoff rods~\cite{bretl2014}, Coserrat rods~\cite{lang2011}, and geometrically exact dynamics splines~\cite{geometricallyexact2008}.
Despite the reduced complexity, these methods still require significant amounts of time to simulate the behavior of the \gls{dlo}, especially for more prolonged movements crucial for efficient motion planning. Moreover, all of the above-mentioned methods require notable knowledge about the manipulated object, like mesh, Young's and shear modulus, or 
mass, but also accurate perceptions systems able to identify all elements of the complex state of the object representation, like twist along the \gls{dlo}.
Finally, due to the idealistic assumptions about the \gls{dlo} model, environmental conditions, and perception system, they may fail in accurately predicting the \gls{dlo} behavior due to reality-gap.


\subsection{Learning-based DLO models}
To address the issues of the model-based approaches mentioned above, data-driven \gls{dlo} models were introduced. They have received much attention in recent years thanks to their flexibility and relatively low computational effort. One of the first approaches to learning \gls{dlo} modeling focuses on learning implicit models.
In~\cite{tamar2019rss}, a Causal InfoGAN is trained to generate plausible transitions in the observations space based on the transitions between latent states, which makes it possible to guide visual servoing. In turn, authors of~\cite{wenbo2021linearlatentdynamics} propose to learn a linear dynamics model in the latent space jointly with an encoder-decoder to encode the \gls{dlo} states and robot actions. In contrast to~\cite{tamar2019rss}, this approach allows directly generating actions that should lead to the goal state. However, it still operates in the image space, which seems suboptimal in terms of generalization.

The learning of \gls{dlo} model based on the low-dimensional \gls{dlo} state representation is presented in~\cite{mitrano2021trust}. This work uses a one-step transition model based on the initial and end state of the \gls{dlo}, represented by the sequence of points and the applied change of grippers' positions. 
In~\cite{mitrano2021trust}, the transition model is based on \gls{mlp}, but without taking advantage of the known structure of the \gls{dlo}. To exploit this, authors of~\cite{bohg2020prediction} used a bidirectional LSTM neural network to model the evolution of the \gls{dlo} state given the applied action. Another approach that takes advantage of representing a \gls{dlo} as a sequence of points is presented in~\cite{wang2022dlomodelgnn}, where a graph neural network based \gls{dlo} model was proposed.
However, the best performance of \gls{dlo} modeling was achieved by the fusion of these approaches -- \gls{inbilstm}~\cite{stoyanov2022}.

Inspired by the jacobian-based \gls{dlo} model~\cite{berenson2013jacobian}, authors of~\cite{mingrui2023tro} trained a neural network to generate jacobian between the grippers and \gls{dlo} velocity based on the actual \gls{dlo} state that locally approximates the \gls{dlo} behavior. By doing so, they imposed a strong local linear prior on the \gls{dlo} model, which led to the state-of-the-art performance in predicting the movement of the simulated \gls{dlo}.

Our paper considers a similar task to the one described in~\cite{mingrui2023tro}. However, while in~\cite{mingrui2023tro} and \cite{wang2022dlomodelgnn} an online adaptation of the trained \gls{dlo} model was proposed, in our work, we focus only on the performance of the so-called \textit{offline models} learned only on the already collected data. We propose an alternative architecture for \gls{dlo} model -- Transformer, which can learn the structure of interactions between points that represent \gls{dlo} induced by the movement of the robot's grippers. Finally, we propose a data augmentation technique that induces a similar bias on the trained model as the one introduced by the jacobian in~\cite{mingrui2023tro} but is not restricted by the assumption of the localness of the deformation.

\section{Neural network-based quasi-static model of a DLO}
The goal of this work is to learn from data a model $f$ for predicting the next \gls{dlo} state $\state_{\idx+1} = f(\state_{\idx}, \eepose_{\idx}, \action_{\idx})$, given the actual \gls{dlo} state $\state_{\idx}$, an actual $\eepose_{\idx}$ pose of the robots \glspl{ee} holding the \gls{dlo}, and a movement of the robots $\action_{\idx}$. 

We define the state of \gls{dlo} as a sequence of $\numPts$ points in 3D space $\{(x_1, y_1, z_1), (x_2, y_2, z_2), \ldots ,(x_{\numPts}, y_{\numPts}, z_{\numPts})\} \in \reals^{\numPts \times 3}$.
The pose of the robots \glspl{ee} consists of the positions $\trans$ and orientations $\rot$ of the left and right robotic arm TCPs i.e. $\eepose = (\trans^L, \rot^L, \trans^R, \rot^R)$, where $L$ and $R$ superscripts denotes the arm. 
We consider our approach quasi-static as we limit our analysis to the states in which \gls{dlo} is not moving and consider only the changes of the steady states, i.e., states with minimal energy of the \gls{dlo} due to the reconfiguration of the end-effectors that hold it. While it is possible to analyze the dynamics of this reconfiguration, we observe that in typical bimanual manipulation of the \gls{dlo} it is often enough to consider the steady states~\cite{bretl2014}, as typically the performed moves are slow. Thus, \gls{dlo} can achieve a steady state almost immediately, especially in the case of stiff \glspl{dlo}. 

A crucial aspect of the problem we consider is that we limit the perception system to a single RGBD camera, and we assume markerless and textureless \gls{dlo}, which distinguishes us from the most state-of-the-art approaches~\cite{stoyanov2022, mingrui2023tro, fiducial}. This assumption is motivated by the fact that in typical bimanual robotic manipulation settings, we cannot stick any markers to the object we want to manipulate and that most of the \glspl{dlo} have no detectable visual or geometrical features that could enable us to track the twist along the \gls{dlo}.

\subsection{Data representations}
\label{sec:data_representations}
To efficiently learn a model, appropriate data representations are essential. We must define how to represent the state of a \gls{dlo} and the orientation of the \glspl{ee}.
Regarding the representation of orientation, we analyzed three alternatives, i.e., quaternions, rotation matrix, and axis angle. 

A more challenging problem is how to represent the state of a \gls{dlo}. We mentioned before that we want to represent the shape of a \gls{dlo} as a sequence of 3D points. However, this does not fully describe the \gls{dlo} state. Besides the shape, to represent the \gls{dlo} state, we should also include the twist along the \gls{dlo}. In general settings, we consider it, but in markerless and textureless settings, it is impossible to detect the twist. Therefore, the system we observe is, in general, partially observable. However, we expect that some notion of the twist can be seen in the geometry of a given \gls{dlo}, having enough data to differentiate between the deformations innate for a given \gls{dlo} instance and the twist along it.

The representation based on 3D points has a significant flaw: the lack of translation invariance, i.e., the representations of the \gls{dlo} of the same shape but located at different positions will differ significantly. To avoid this, we represent the \gls{dlo} as a sequence of points described in the coordinate system located in the middle of the right gripper pad and orientation aligned with the coordinate system of the right manipulator base. This introduces translation invariance and, at the same time, maintains the direction of the gravity vector w.r.t. the \gls{dlo}. 
Another approach to achieve the translation invariance of the \gls{dlo} state is to represent it as a sequence of difference vectors between points, i.e., a sequence of edges.

Another decision one has to make is how to represent the actual positions of the \glspl{ee} $\trans^L_i, \trans^R_i$ and their movements $\action_i$. 
A similar transformation as for the \gls{dlo} state can be done for the gripper positions. We can omit the position of the right gripper and express the position of the left one in the right gripper coordinate system.
While the move of the grippers $\action_i$ can be represented by the end pose of the left arm and orientation of the right one $\action_i = (\trans^L_{i+1}, \rot^L_{i+1}, \rot^R_{i+1})$, or by the difference between the end and initial states of the grippers $\action_i = \pare{\trans^L_{i+1} - \trans^L_i, \pare{\rot^L_{i}}^{-1}\rot^L_{i+1}, \pare{\rot^R_{i}}^{-1}\rot^R_{i+1}}$.

\vspace{0.1cm}
\begin{figure*}[t]
    \includegraphics[width=\linewidth]{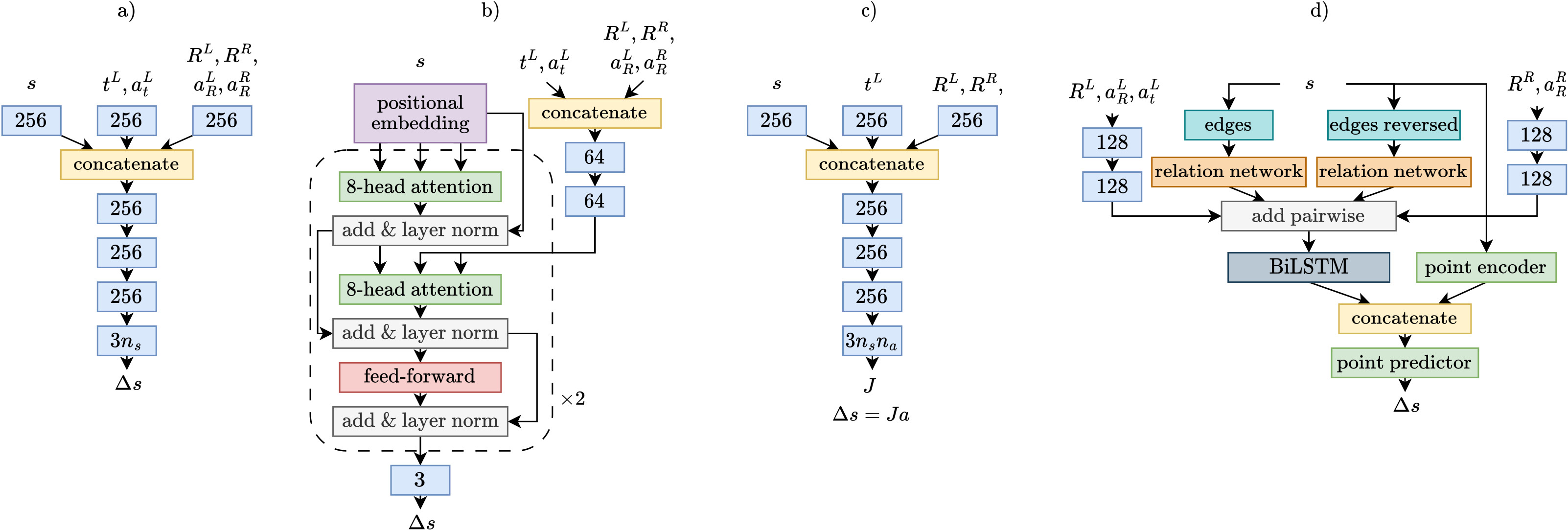}
    \vspace{-0.5cm}
    \caption{Architectures of the neural network-based \gls{dlo} models used in experiments: a) \gls{mlp}, b) Transformer, c) JacMLP~\cite{mingrui2023tro} nad d) \gls{inbilstm}~\cite{stoyanov2022}.}
    \label{fig:architectures}
\vspace{-0.5cm}
\end{figure*}

\subsection{Neural network architectures}
Having the data representations defined, we can focus on how to process them to predict the \gls{dlo} motion accurately.
We propose not to approximate the function $f$ directly but instead use a neural network to approximate the change of the \gls{dlo} state $\Delta\state_{\idx} = f(\state_{\idx}, \eepose_{\idx}, \action_{\idx}) - \state_{\idx}$ after applying a move of the grippers $\action_{\idx}$, given the actual state of the \gls{dlo}$\state_{\idx}$ and grippers $\eepose_{\idx}$.
In this paper, we considered four neural network architectures: (i) a simple \gls{mlp}, (ii) our proposed Transformer-based network~\cite{transformer}, and two baselines (iii) adjusted \gls{inbilstm} architecture~\cite{stoyanov2022}, and (iv) inspired by~\cite{mingrui2023tro} \gls{mlp} that predicts the jacobian of the \gls{dlo} w.r.t. grippers. All considered architectures are shown in Figure~\ref{fig:architectures}.

In Figure~\ref{fig:architectures}a, a simple \gls{mlp} is shown. It generates embeddings of the \gls{dlo} state $\state$, positional information about the left gripper $(\trans^{L}, \action^{L}_{\trans})$, where $\action^{L}_{\trans}$ denotes the translational part of the left gripper movement, and rotational components of the initial state of the grippers $\rot^{L}, \rot^{R}$ as well as the rotational components of the movements of both arms $\action^{L}_{\rot}, \action^{R}_{\rot}$. Then, concatenated embeddings are processed with a sequence of fully-connected layers, and finally, the estimation of the $\Delta\state$ is generated.

Figure~\ref{fig:architectures}b illustrates our proposed approach, a Transformer-based architecture. It is inspired by the attention mechanism~\cite{transformer}, which is supposed to capture better the interactions between the pose and movement of grippers, which serve as the context, and parts of the \gls{dlo} on which we apply the attention mechanism. While the whole architecture is based on the Transformer from~\cite{transformer}, we (i) have not introduced any masking of the cable state signal, (ii) simplified the processing of the context, which in our case is not a sequence, and, because our considered task is a regression, not a classification, (iii) we have not used softmax activation at the final processing step.

In Figure~\ref{fig:architectures}c baseline model is shown, noted as JacMLP, whose architecture is similar to the \gls{mlp}. However, by drawing inspiration from the recent work~\cite{mingrui2023tro}, it is predicting not the \gls{dlo} displacement, but the jacobian matrix $\jac$, representing the local linear transformation $\Delta\state = \jac \action$ between the change of the grippers poses, denoted by $\action$, and the change of the \gls{dlo} state $\Delta\state$. We also tested the RBFN model proposed in~\cite{mingrui2023tro} but obtained inferior performance.

Figure~\ref{fig:architectures}d illustrates the second baseline an \gls{inbilstm} architecture proposed in \cite{stoyanov2022}. We adjusted it to comply with the considered task. To do so, we preprocessed poses and movements of both \glspl{ee} and added their latent representations to the latent representation of the DLO's outermost edges.
The rest of the architecture remained the same as in \cite{stoyanov2022}, with slightly reduced dimensionality of the latent space from $150$ to $128$.

\section{Dataset}

\subsection{Dataset collection}

To collect the dataset, we used two UR3 robots with custom 3D printed grippers, Microsoft Azure Kinect camera, and our DLOFTBs algorithm~\cite{dloftbs} based on B-splines for fast tracking of \glspl{dlo}. Data were collected in sequences of 20 random arms moves with enforced constraints to prevent ripping off the \gls{dlo}. For each sequence, the initial states of the robots and \gls{dlo} were manually chosen to cover a broad spectrum of system configurations. Next, we removed samples with visible issues with depth measurements, which can happen for very thin objects. Finally, we generated data points by merging any pair of system configurations from the same sequence. As a result, each sample from the dataset consists of the initial and end \glspl{ee} poses $\eepose_{\idx-1}, \eepose_\idx$ and \gls{dlo} states $\state_{\idx-1}, \state_\idx$.

For detecting the \gls{dlo} shape on the RGBD image, we used a DLOFTBs algorithm~\cite{dloftbs} acting on the \glspl{dlo} masks extracted based on a hue-based segmentation. The output of this method is a 3D B-spline curve representing the shape of the \gls{dlo}. However, we observed that in a bimanual \gls{dlo} manipulation setting, the ends of the \gls{dlo} are often not clearly visible to the camera. Thus, the length of the detected \gls{dlo} is changing. This makes it harder to stably track the points on the \gls{dlo}. Therefore, we propose utilizing the information about the grippers handling a DLO and including their TCPs in the list of points on the \gls{dlo}. Nevertheless, to achieve stable representation, we want to use a sequence of points that are equally distant from each other. Thus, we decided to fit a B-spline to the points on the \gls{dlo} and grippers TCPs, and then compute $N$ equally distant points on the \gls{dlo} (where the distance is computed along the B-spline curve).
Moreover, we observed that using the Kinect Azure sensor, the quality of the depth estimation of the \gls{dlo} is reduced when the depth of the background is close to the \gls{dlo}, which often happens at the end of the \gls{dlo}, as they are close to the grippers. This might be caused by the internal Kinect algorithms smoothing for surface estimation. The low number of depth points at DLO are regarded as noise or outliers against most points constituting the background's surface. To address this issue, we decided to neglect the depth of the points on ends of a \gls{dlo} that lie too close to the grippers and rely on the B-spline interpolation.
\noindent In this way, we collected several datasets:
\begin{itemize}[noitemsep,topsep=-\parskip]
    \item \SI{50}{\centi\metre} of two-wire cable (38698 training / \\ 9182 validation / 6200 test samples),
    \item \SI{45}{\centi\metre} of two-wire cable (8848/3284/2474),
    \item \SI{40}{\centi\metre} of two-wire cable (7414/2824/3026),
    \item \SI{50}{\centi\metre} of solar cable (3378/708/1258),
    \item \SI{50}{\centi\metre} of braided cable (3674/1042/1420).
\end{itemize}
To collect them, we used 3 different cables: two-wire, solar (used in photovoltaic installations), and braided, which differ notably in terms of stiffness and plasticity and have comparable diameters of 6.9mm, 5.8mm, and 5.8mm, respectively. In terms of differences, the stiffest is the solar cable, the medium is two-wire, and the less stiff is the braided one. Braided cable resembles a slightly stiffened rope and is characterized by the highest plasticity. In turn, the plasticity of solar and two-wire cables is comparable. However, the solar one retains deformation more strongly.


\subsection{Data augmentation}

In this paper, we propose a simple yet efficient data augmentation technique that can be applied to introduce an essential inductive bias to the predictions of the neural network-based \gls{dlo} model.
We assumed that the considered \gls{dlo} motions are quasi-static, and we would like to model them using a function $f(\state_{\idx}, \eepose_{\idx}, \action_{\idx})$. In this setting, if grippers do not move and no variable external forces act on the \gls{dlo}, we expect the \gls{dlo} to remain in the same state. Therefore, we would like a neural network-based model to exhibit the same property, i.e., $f(\state_{\idx}, \eepose_{\idx}, \action_{\idx}) = 0$ if $\action_{\idx}$ is equal to actual \gls{ee} pose $\eepose_{\idx}$ in case of the positional representation or $\action_{\idx}$ is equal to identity in case of representing it as a difference.
Because this property is not automatically satisfied due to the architecture of our models, except the MLPJac, which in particular can satisfy it if the action representation for no move is equal to zero vector, we propose to add to the dataset samples that represent the case of no motion of the arms at all \gls{dlo} configurations from the dataset. 
In the experimental section, we show that this simple trick substantially improves the accuracy of the predictions of all considered models except MLPJac.

\subsection{Test-time \gls{dlo} scaling}
\label{sec:scaling}

One of the typical limitations of the neural network-based models of the \glspl{dlo} is their lack of adaptation to the easy-to-get \gls{dlo} parameters such as length. Typically, it is assumed that the network is trained to work only for a given \gls{dlo} type and length. This problem was already addressed in~\cite{mingrui2023tro}, however, only in the context of predicting the local linear model of the \gls{dlo} by scaling the parts of the Jacobian matrix. Instead, we propose to include the information about length more generally, i.e., in the input to the neural network. Particularly to consider a neural network trained on the \gls{dlo} of length $l_1$. To make a test data distribution of the same \gls{dlo} but of length $l_2$ similar to the one on which it was trained, we propose to scale the \gls{dlo} representation and positions of the left arm by the factor of $\frac{l_1}{l_2}$. Of course, this type of scaling does not correspond to actual changes in cable behavior caused by the length change, although it is a simple heuristic meant to approximate it. We conjecture that this scaling may be more effective for small changes in the length of the \gls{dlo}, as for the bigger ones, the impact of the grippers rotations that is not scaling linearly with length may be predominant.

\section{Experimental evaluation}
The results section will first focus on our proposed Transformer-based model. Then, we will show gains stemming from the proposed data augmentation method. We will compare our results against state-of-the-art approaches. Further, we elaborate more on the properties of the proposed approaches—namely, generalization and inference time. Finally, we will show the performance of the proposed method in model-based DLO bimanual manipulation. 
\subsection{DLO shape prediction}
\subsubsection{Baseline experiment}
\begin{figure}[t]
    \includegraphics[width=\linewidth]{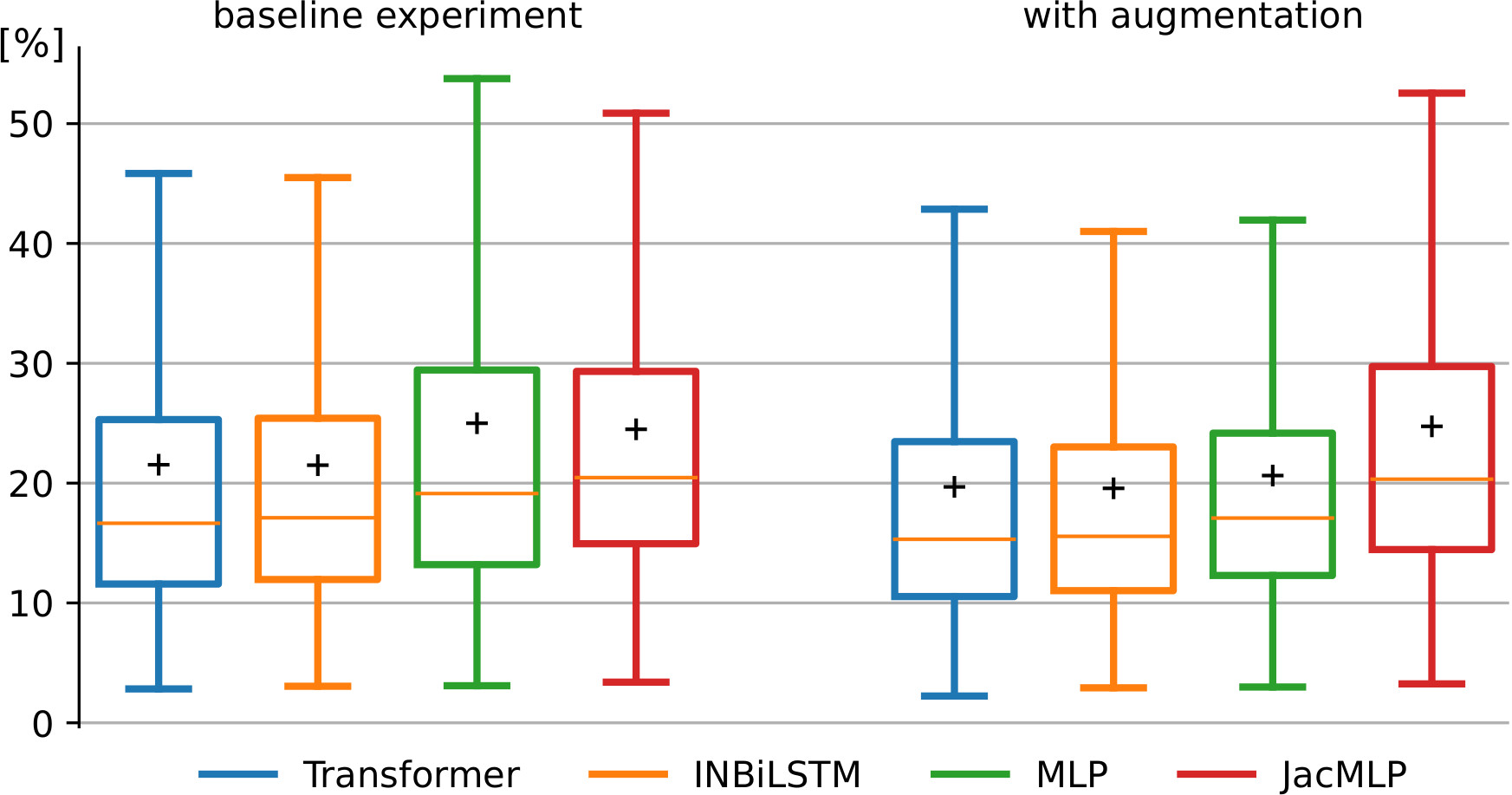}
    \caption{Relative prediction error [\%] of the neural DLO models on the 50cm long two-wire cable test set. In the left part of the plot are the baseline experiment results, while in the right one, we present the effects of applying our proposed data augmentation technique. Improvement is visible for all models except JacMLP.
    }
    \label{fig:stats}
\end{figure}

We start our experiments by comparing all considered learning-based \gls{dlo} models trained on the dataset of 50cm long two-wire cable. The results obtained by these models on the test subset of the two-wire cable dataset are shown in the left part of Figure~\ref{fig:stats}, denoted as \textit{baseline}. As a metric, we use the ratio of $\mathcal{L}_3$ error introduced in~\cite{dloftbs} to the $\mathcal{L}_3$ error computed between initial and ground-truth cable pose after the move to account for the variable scale of deformations in the dataset. 

One can see that the proposed Transformer architecture obtains the best result. However, almost the same prediction accuracy was achieved by the \gls{inbilstm}. Significantly higher errors are achieved by the \gls{mlp} and JacMLP, which fared the worst.
While we performed the tests with all the data representations introduced in Section~\ref{sec:data_representations}, we found out that the impact of the representation type is minor. Thus, for the sake of clarity of the presentation, we report the test set statistics only for the best of them. Particularly, for the Transformer architecture, the best results were obtained for representing the orientation using axis-angle, move by the difference of poses, and \gls{dlo} by edges, for \gls{inbilstm} orientation -- axis-angle, move -- end-pose and \gls{dlo} -- a sequence of points, for \gls{mlp} with orientation -- rotation matrix, move -- end-pose and \gls{dlo} -- a sequence of points, while for JacMLP with orientation -- rotation matrix, move --  a difference of poses, and \gls{dlo} -- a sequence of edges.
 
\subsubsection{Data augmentation}
\begin{figure}[t]
    \includegraphics[width=\linewidth]{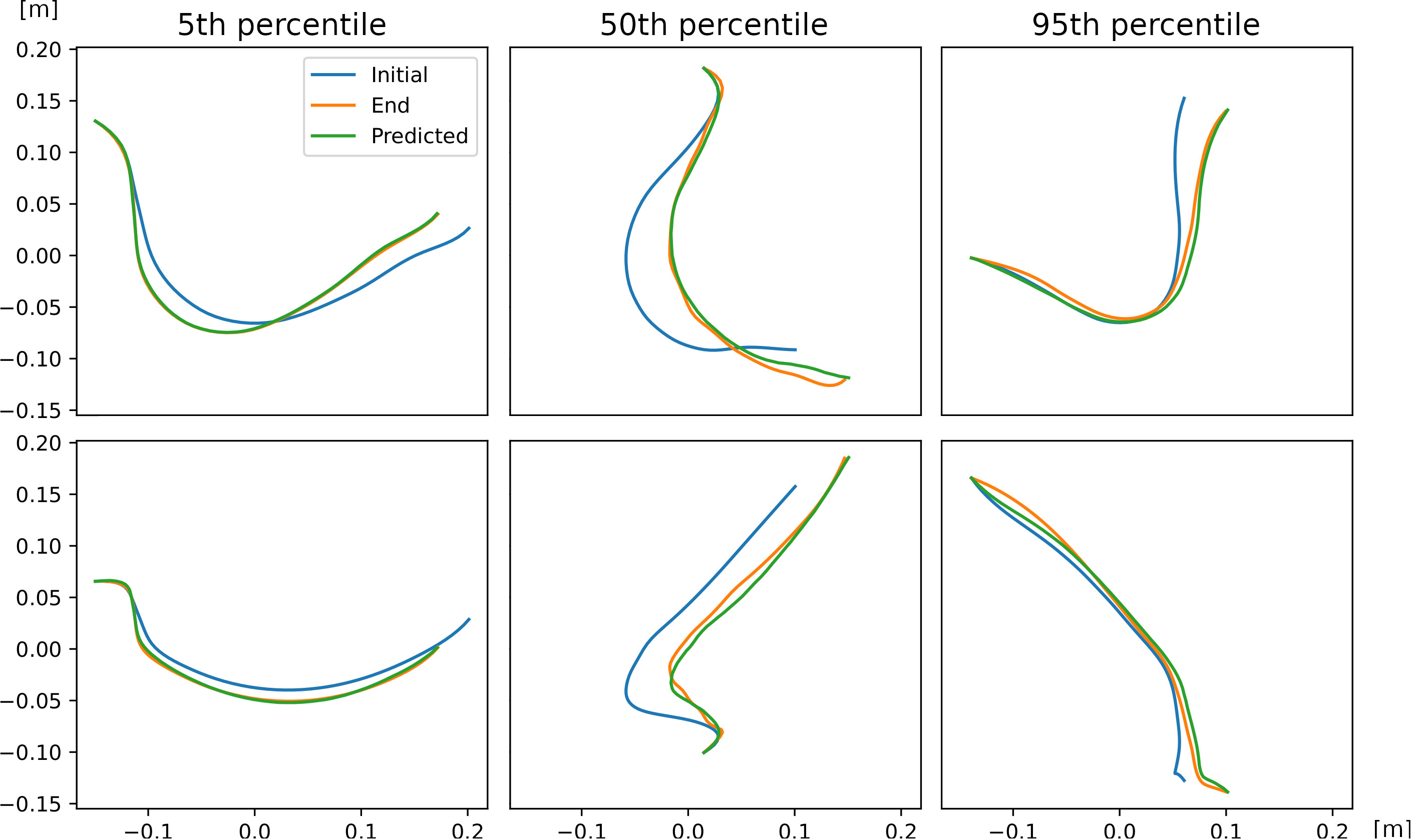}
    \caption{Samples from the test set with predictions made with \gls{mlp} model, which are in 5th, 50th, and 95th percentile of errors. The top row presents the front view from the camera, while the bottom one is the view from the top.}
    \label{fig:examples}
\end{figure}

In this part of the experiment, we assess the impact of the proposed data augmentation technique on the results achieved by the \gls{dlo} models in the previous experiment and present obtained results in the right part of Figure~\ref{fig:stats}.
One can observe that simple \gls{mlp}, whose performance was significantly dominated in the previous experiment, thanks to the use of data augmentation, achieves accuracy comparable to the best models.
Moreover, augmentation improved the performance of all considered models except the JacMLP, on average by 15\% for \gls{mlp} and by 6\% for \gls{inbilstm} and Transformer. Nevertheless, the median errors at the level of 15.31\% for Transformer, 15.56\%, 17.08\% for \gls{mlp}, and 20.33\% for JacMLP, may seem relatively big. To better visualize what kind of errors these values represent, we show in Figure~\ref{fig:examples} examples of the \gls{dlo} behavior predictions made by the \gls{mlp} model with augmentation that is in 5th, 50th, and 95th error percentile.

\subsection{Generalization}

\subsubsection{DLO length}
\begin{figure}[t]
    \includegraphics[width=\linewidth]{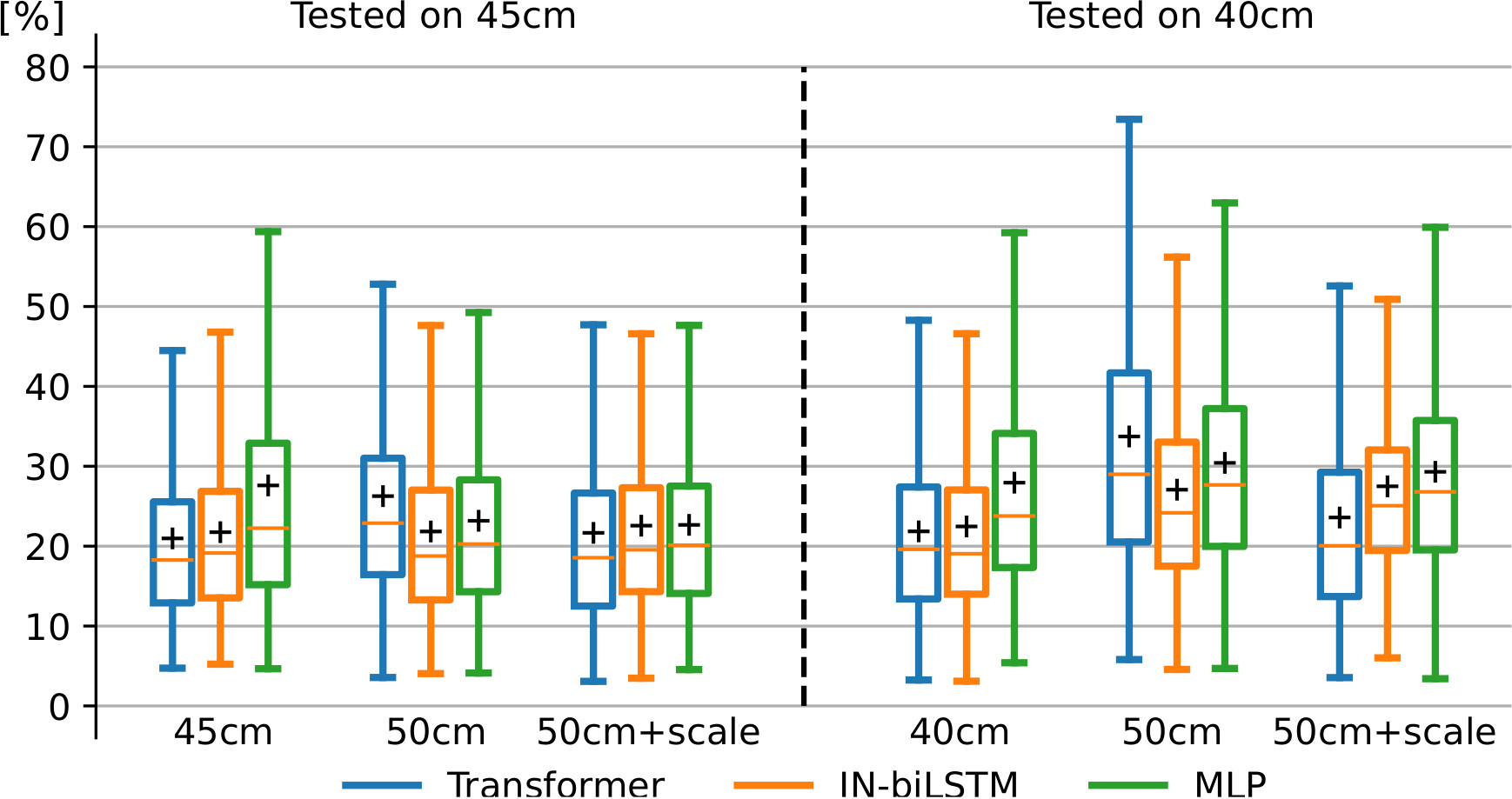}
    \caption{Relative prediction errors [\%] for the task of predicting the movement of the \gls{dlo} of a different length. The Transformer-based model is able to achieve the same performance as the baseline (trained on the test length \gls{dlo}) if the appropriate scaling is applied (x-axis training set).}
    \label{fig:dlo_length}
\end{figure}

One of the commonly overlooked generalizations regarding the neural network-based \gls{dlo} modeling is the length of the cable being modeled. 
In this experiment, we assess the performance of all architectures that achieved similar performance in the previous experiment on the task of predicting the movement of the same \gls{dlo} but shorter, i.e., 45cm and 40cm. To have a reference point, we referred these results to the ones obtained by the models trained on the datasets with shorter \gls{dlo}. Moreover, we test the impact of our proposed scaling procedure (see Section~\ref{sec:scaling}).
Results of this comparison are presented in~Figure~\ref{fig:dlo_length}. 
One can see, that if the difference in length is relatively small (10\%), all of the considered architectures achieve results comparable to the baseline. However, increasing the length difference to 10cm causes notable degradation of the performance of the models trained on the dataset of 50cm long cable. 
While in both cases, the worst-performing model was the Transformer, if the proposed scaling procedure is applied, we observe a significant improvement and the performance at the same level as for this model trained on the target dataset.
Interestingly, the same scaling procedure results in no improvement or even decrease in the performance of the \gls{mlp} and \gls{inbilstm} models, which may suggest that only Transformer was able to capture the right intuition about the nature of the modeled system.

\subsubsection{Different DLOs}
\begin{figure}[t]
    \includegraphics[width=\linewidth]{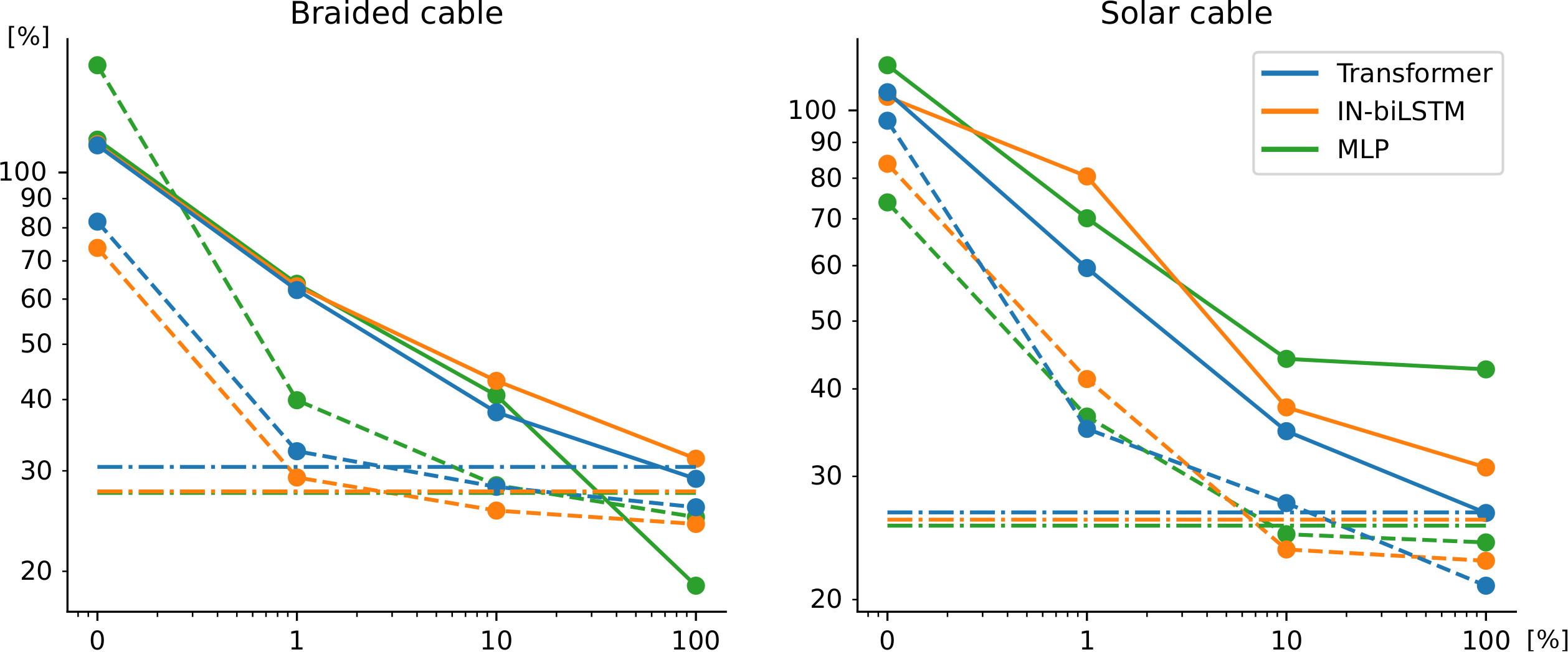}
    \caption{Analysis of the generalization to different \gls{dlo} types based on the relative error [\%] in a function of the part of the data set [\%] of the target \gls{dlo}. The solid line represents the performance of the models trained from scratch of the target dataset, dashed trained on the target but pretrained on the two-wire cable, while the dashed-dot line is the baseline model which was trained only on the two-wire cable dataset.}
    \label{fig:dlo_type}
\end{figure}

Another very important ability of the model is to generalize to previously unseen \glspl{dlo} or at least to efficiently learn the model of the new ones considering already gathered knowledge by pretraining.
We evaluated these capabilities on 2 different \glspl{dlo}: (i) braided cable and (ii) solar cable. We compared the performance of the models trained on the two-wire cable dataset with those trained from scratch on the braided and solar cables datasets and those that were first pre-trained on two-wire and then trained on braided and solar.
To check how much data from a new cable is needed to train a new model efficiently, we performed these tests using 100\%, 10\%, 1\%, and 0.1\% of the target dataset. The results of these experiments are presented in Figure~\ref{fig:dlo_type}.
The baseline models that were trained only on the two-wire cable data achieved decent prediction performance. To achieve an improvement w.r.t. them, one needs to start from a pretrained model and use at least hundreds of the datapoints with the target \gls{dlo}. It is clearly visible that in almost all cases, the knowledge of the behavior of one cable gives an advantage in learning the behavior of the new one. However, for a braided cable, it turned out that given enough training data, the pretraining may limit the capability to learn.
Nevertheless, pretraining not only, in general, improves prediction accuracy but also enables models to be trained more than 100 times faster.
Finally, we do not see any significant evidence that one of the compared architectures is better for generalizing the new \gls{dlo} types. However, the Transformer in both cases was slightly worse when it did not have access to the target dataset.

\subsection{Infernece time}
\begin{figure}[t]
    \includegraphics[width=\linewidth]{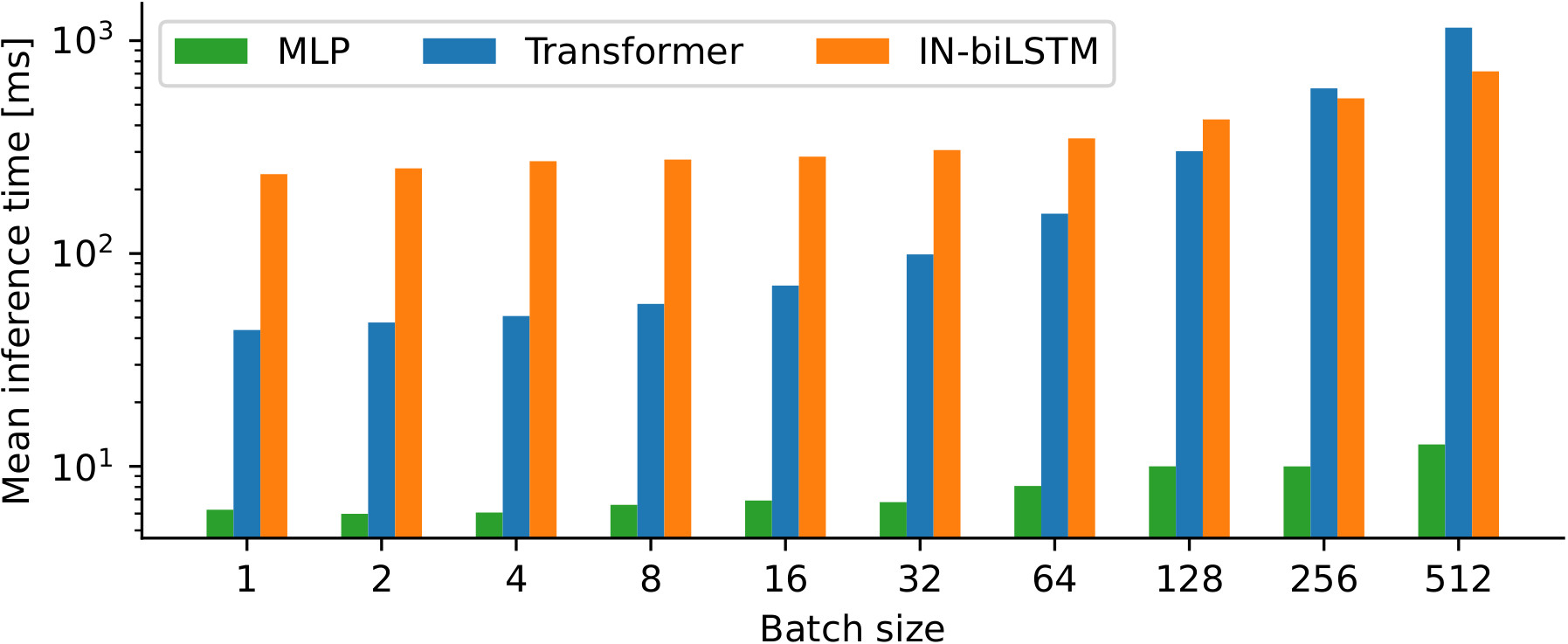}
    \caption{Inference times of the \gls{dlo} models.}
    \label{fig:timings}
\end{figure}

Another important parameter of the machine learning models, except for their accuracy, is the inference time. This is particularly important because these models can become a part of the planning or control pipeline, which typically has a limited time budget.
We compared the timings of the best \gls{mlp}, \gls{inbilstm}, and Transformer models using Intel(R) Core(TM) i7-9750H CPU with 32GB RAM and presented results in Figure~\ref{fig:timings}. One can see that the Transformer is faster than \gls{inbilstm}. However, for the growing batch size, the difference reduces, and starting from batch size equal to 256 \gls{inbilstm} becomes faster.
Nevertheless, the best performance is achieved by a simple \gls{mlp}, which outperforms both Transformer and \gls{inbilstm} models by a large margin, as it is more than 50 times faster than \gls{inbilstm} and 7 times faster than Transformer.

\subsection{Model-based DLO bimanual manipulation}
In the last experiment, our goal is to evaluate the possibility of using the trained froward \gls{dlo} model to compute its inverse, i.e., find a movement of grippers that results in the desired deformation and movement of the \gls{dlo}. While this may be achieved in several ways, we used one of the simplest yet efficient approaches, i.e., \gls{cem}~\cite{rubinstein2004cem}. To perform this experiment, we choose the \gls{mlp} model as, thanks to the proposed data augmentation method, it achieved a decent prediction accuracy, and it can be evaluated very fast. The short inference time of the model is crucial for making the planning of the movements of the grippers fast, which is desirable in potential industrial applications.

The core idea of \gls{cem} in our setting is to:
(i) draw random gripper movements from some randomly initialized normal distribution, (ii) predict the state of the \gls{dlo} after applying these movements using our neural \gls{dlo} model, (iii) assess the predicted shapes of the \gls{dlo} in reference to the desired one using the mean absolute difference between points on \gls{dlo}, (iv) update the normal distribution parameters based on several actions that resulted in smallest shapes differences -- elite actions, (v) repeat the process until convergence or exceding the maximum number of iterations. In our experiments, we set the number of drawn samples to 64, the number of elite actions to 8, and the maximum number of iterations to 10.
With these settings, the whole planning process takes less than 100ms on the same machine as in the previous section.

\begin{figure*}[t]
    \includegraphics[width=\linewidth]{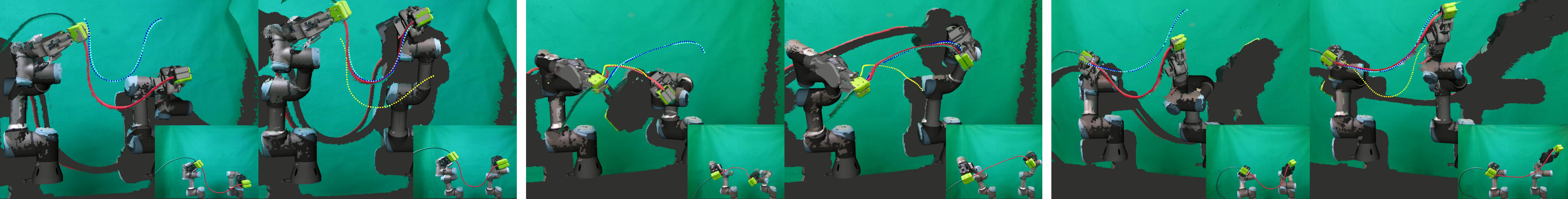}
    \caption{Examples of the bi-manual \gls{dlo} shaping using neural-network-based \gls{dlo} model and \gls{cem}. Each scenario consists of an initial frame (red) and a final frame (green), with the predicted and desired shapes marked with light and dark blue, and the initial detected shape marked with yellow.}
    \label{fig:planning}
\end{figure*}

Sample results of this experiment are presented on the frames in Figure~\ref{fig:planning}, while the results of all performed manipulation trials are presented in the video attachment. One can see that the proposed model allowed for a quite accurate prediction of the gripper's positions and orientations that results in obtaining a similar 3D shape of the \gls{dlo} to the desired one. These results were obtained on a wide range of reshaping problems like bending, unbending, and large deformations in the direction parallel to the camera's optical axis. Obtained shape reconstruction errors are, on average, about 1.5cm according to the $\mathcal{L}_3$ metric proposed in~\cite{dloftbs}.

\section{Conclusions}
In this paper, we analyzed the problem of learning a quasi-static 3D model of a markerless \gls{dlo} manipulated with two robotic arms. We proposed a Transformer architecture that outperforms the State-of-the-Art in terms of prediction accuracy, and we introduced a data augmentation technique that improves the prediction performance even more for almost all considered neural network architectures. By using this method, a simple \gls{mlp} was able to achieve results comparable to Transformer in terms of accuracy while being more than 7 times faster. Moreover, we analyzed the ability of the considered models to generalize to different \gls{dlo} types and lengths. We showed that a model pretrained on the different cable types was able to predict the behavior of the new cable quite accurately, even without training on the target one. Moreover, using a pretrained model that was then trained on the dataset with target cable in most of the cases outperforms the solution trained from scratch. 
Regarding the \gls{dlo} length, we showed that if the length of the \gls{dlo} notably
changed between training and test time, the performance of the learning-based models decreased. However, if the proposed scaling procedure is applied to the Transformer model, then the achieved performance is nearly the same as in the baseline model trained on the target dataset.
Finally, we evaluated the fast \gls{mlp}-based \gls{dlo} model trained on the augmented data in the task of predicting the movement of the grippers to obtain a desired shape of the \gls{dlo} and demonstrated its speed and accuracy.





\bibliographystyle{IEEEtran}
\bibliography{IEEEexample}

\begin{thebibliography}{10}
\providecommand{\url}[1]{#1}
\csname url@rmstyle\endcsname
\providecommand{\newblock}{\relax}
\providecommand{\bibinfo}[2]{#2}
\providecommand\BIBentrySTDinterwordspacing{\spaceskip=0pt\relax}
\providecommand\BIBentryALTinterwordstretchfactor{4}
\providecommand\BIBentryALTinterwordspacing{\spaceskip=\fontdimen2\font plus
\BIBentryALTinterwordstretchfactor\fontdimen3\font minus
  \fontdimen4\font\relax}
\providecommand\BIBforeignlanguage[2]{{%
\expandafter\ifx\csname l@#1\endcsname\relax
\typeout{** WARNING: IEEEtran.bst: No hyphenation pattern has been}%
\typeout{** loaded for the language `#1'. Using the pattern for}%
\typeout{** the default language instead.}%
\else
\language=\csname l@#1\endcsname
\fi
#2}}

\bibitem{wlatersson2022routing}
G.~A. Waltersson, R.~Laezza, and Y.~Karayiannidis, ``Planning and control for
  cable-routing with dual-arm robot,'' in \emph{2022 Int. Conf. on Robotics and
  Automation (ICRA)}, 2022, pp. 1046--1052.

\bibitem{tomizuka2022routing}
S.~Jin, W.~Lian, C.~Wang, M.~Tomizuka, and S.~Schaal, ``Robotic cable routing
  with spatial representation,'' \emph{IEEE Robotics Automation Lett.}, vol.~7,
  no.~2, pp. 5687--5694, 2022.

\bibitem{knots}
M.~Saha and P.~Isto, ``Manipulation planning for deformable linear objects,''
  \emph{IEEE Trans. on Robotics}, vol.~23, no.~6, pp. 1141--1150, 2007.

\bibitem{priya2021untangling}
P.~Sundaresan, J.~Grannen, B.~Thananjeyan, A.~Balakrishna, J.~Ichnowski,
  E.~Novoseller, M.~Hwang, M.~Laskey, J.~Gonzalez, and K.~Goldberg,
  ``{Untangling Dense Non-Planar Knots by Learning Manipulation Features and
  Recovery Policies},'' in \emph{Proc. of Robotics: Sci. and Systems}, Virtual,
  July 2021.

\bibitem{tomizuka2021optimization}
S.~Jin, D.~Romeres, A.~Ragunathan, D.~K. Jha, and M.~Tomizuka, ``Trajectory
  optimization for manipulation of deformable objects: Assembly of belt drive
  units,'' in \emph{2021 IEEE Int. Conf. on Robotics and Automation (ICRA)},
  2021, pp. 10\,002--10\,008.

\bibitem{zhu2020contacts}
J.~Zhu, B.~Navarro, R.~Passama, P.~Fraisse, A.~Crosnier, and A.~Cherubini,
  ``Robotic manipulation planning for shaping deformable linear objects
  withenvironmental contacts,'' \emph{IEEE Robotics Automation Lett.}, vol.~5,
  no.~1, pp. 16--23, 2020.

\bibitem{wiring_harness_assembly}
P.~Kicki, M.~Bednarek, P.~Lembicz, G.~Mierzwiak, A.~Szymko, M.~Kraft, and
  K.~Walas, ``Tell me, what do you see?—interpretable classification of
  wiring harness branches with deep neural networks,'' \emph{Sensors}, vol.~21,
  no.~13, 2021.

\bibitem{berenson2015tightholes}
W.~Wang, D.~Berenson, and D.~Balkcom, ``An online method for tight-tolerance
  insertion tasks for string and rope,'' in \emph{2015 IEEE Int. Conf. on
  Robotics and Automation (ICRA)}, 2015, pp. 2488--2495.

\bibitem{surgical_sutures}
S.~Sen, A.~Garg, D.~V. Gealy, S.~McKinley, Y.~Jen, and K.~Goldberg,
  ``Automating multi-throw multilateral surgical suturing with a mechanical
  needle guide and sequential convex optimization,'' in \emph{2016 IEEE Int.
  Conf. on Robotics and Aut. (ICRA)}, 2016, pp. 4178--4185.

\bibitem{FEM}
J.~{Sanchez}, C.~M. {Mateo}, J.~A. {Corrales}, B.~{Bouzgarrou}, and
  Y.~{Mezouar}, ``Online shape estimation based on tactile sensing and
  deformation modeling for robot manipulation,'' in \emph{2018 IEEE/RSJ Int.
  Conf. on Intelligent Robotics and Systems (IROS)}, 2018, pp. 504--511.

\bibitem{apetit_siciliano}
A.~{Petit}, V.~{Lippiello}, and B.~{Siciliano}, ``Real-time tracking of 3d
  elastic objects with an rgb-d sensor,'' in \emph{2015 IEEE/RSJ Int. Conf. on
  Intelligent Robotics and Systems (IROS)}, 2015, pp. 3914--3921.

\bibitem{coros2018interactive}
S.~Duenser, J.~M. Bern, R.~Poranne, and S.~Coros, ``Interactive robotic
  manipulation of elastic objects,'' in \emph{2018 IEEE/RSJ Int. Conf. on
  Intelligent Robotics and Systems (IROS)}, 2018, pp. 3476--3481.

\bibitem{coros2021dynamic}
S.~Zimmermann, R.~Poranne, and S.~Coros, ``Dynamic manipulation of deformable
  objects with implicit integration,'' \emph{IEEE Robotics Automation Lett.},
  vol.~6, no.~2, pp. 4209--4216, 2021.

\bibitem{cosserat_rod}
J.~Spillmann and M.~Teschner, ``Corde: Cosserat rod elements for the dynamic
  simulation of one-dimensional elastic objects,'' in \emph{Proc. of the 2007
  ACM SIGGRAPH/Eurographics Symp. on Comput. Animation}, 2007, p. 63–72.

\bibitem{lang2011}
H.~Lang, J.~Linn, and M.~Arnold, ``Multi-body dynamics simulation of
  geometrically exact cosserat rods,'' \emph{Multibody System Dynamics},
  vol.~25, no.~3, pp. 285--312, Mar 2011.

\bibitem{kavan2016cosserat}
T.~Kugelstadt and E.~Schömer, ``{Position and Orientation Based Cosserat
  Rods},'' in \emph{Eurographics/ ACM SIGGRAPH Symp. on Comput. Animation},
  2016.

\bibitem{bretl2014}
T.~Bretl and Z.~McCarthy, ``Quasi-static manipulation of a kirchhoff elastic
  rod based on a geometric analysis of equilibrium configurations,'' \emph{The
  Int. J. of Robotics Res.}, vol.~33, no.~1, pp. 48--68, 2014.

\bibitem{geometricallyexact2022}
Y.~Liu, K.~Song, and L.~Meng, ``A geometrically exact discrete elastic rod
  model based on improved discrete curvature,'' \emph{Comput. Methods in
  Applied Mechanics and Engineering}, vol. 392, p. 114640, 2022.

\bibitem{gianluca_ICPS}
G.~{Palli}, ``Model-based manipulation of deformable linear objects by
  multivariate dynamic splines,'' in \emph{2020 IEEE Conf. on Industrial
  Cyberphysical Systems (ICPS)}, vol.~1, 2020, pp. 520--525.

\bibitem{khalifa2022}
A.~Khalifa and G.~Palli, ``New model-based manipulation technique for reshaping
  deformable linear objects,'' \emph{The Int. J. of Adv. Manuf. Technol.}, vol.
  118, no.~11, pp. 3575--3583, Feb 2022.

\bibitem{stoyanov2022}
Y.~Yang, J.~A. Stork, and T.~Stoyanov, ``Learning differentiable dynamics
  models for shape control of deformable linear objects,'' \emph{Robotics and
  Autonomous Systems}, vol. 158, p. 104258, 2022.

\bibitem{mitrano2021trust}
P.~Mitrano, D.~McConachie, and D.~Berenson, ``Learning where to trust
  unreliable models in an unstructured world for deformable object
  manipulation,'' \emph{Sci. Robotics}, vol.~6, no.~54, p. eabd8170, 2021.

\bibitem{mingrui2023tro}
M.~Yu, K.~Lv, H.~Zhong, S.~Song, and X.~Li, ``Global model learning for large
  deformation control of elastic deformable linear objects: An efficient and
  adaptive approach,'' \emph{IEEE Trans. on Robotics}, vol.~39, no.~1, pp.
  417--436, 2023.

\bibitem{wang2022dlomodelgnn}
C.~Wang, Y.~Zhang, X.~Zhang, Z.~Wu, X.~Zhu, S.~Jin, T.~Tang, and M.~Tomizuka,
  ``Offline-online learning of deformation model for cable manipulation with
  graph neural networks,'' \emph{IEEE Robotics Automation Lett.}, vol.~7,
  no.~2, pp. 5544--5551, 2022.

\bibitem{bohg2020prediction}
M.~Yan, Y.~Zhu, N.~Jin, and J.~Bohg, ``Self-supervised learning of state
  estimation for manipulating deformable linear objects,'' \emph{IEEE Robotics
  Automation Lett.}, vol.~5, no.~2, pp. 2372--2379, 2020.

\bibitem{wenbo2021linearlatentdynamics}
W.~Zhang, K.~Schmeckpeper, P.~Chaudhari, and K.~Daniilidis, ``Deformable linear
  object prediction using locally linear latent dynamics,'' in \emph{2021 IEEE
  Int. Conf. on Robotics and Automation (ICRA)}, 2021, pp. 13\,503--13\,509.

\bibitem{dloftbs}
P.~Kicki, A.~Szymko, and K.~Walas, ``{DLOFTBs – Fast Tracking of Deformable
  Linear Objects with B-splines},'' in \emph{2023 IEEE Int. Conf. on Robotics
  and Automation (ICRA)}, 2023, pp. 7104--7110.

\bibitem{transformer}
A.~Vaswani, N.~Shazeer, N.~Parmar, J.~Uszkoreit, L.~Jones, A.~N. Gomez, L.~u.
  Kaiser, and I.~Polosukhin, ``Attention is all you need,'' in \emph{Advances
  in Neural Information Processing Systems}, I.~Guyon, U.~V. Luxburg,
  S.~Bengio, H.~Wallach, R.~Fergus, S.~Vishwanathan, and R.~Garnett, Eds.,
  vol.~30.\hskip 1em plus 0.5em minus 0.4em\relax Curran Associates, Inc.,
  2017.

\bibitem{interaction2016networks}
P.~Battaglia, R.~Pascanu, M.~Lai, D.~J. Rezende, and K.~kavukcuoglu,
  ``Interaction networks for learning about objects, relations and physics,''
  in \emph{Proc. of the 30th Int. Conf. on Neural Information Processing
  Systems}, ser. NIPS'16.\hskip 1em plus 0.5em minus 0.4em\relax Red Hook, NY,
  USA: Curran Associates Inc., 2016, p. 4509–4517.

\bibitem{saha1993rbfn}
A.~Saha, C.-L. Wu, and D.-S. Tang, ``Approximation, dimension reduction, and
  nonconvex optimization using linear superpositions of gaussians,'' \emph{IEEE
  Trans. on Comput.}, vol.~42, no.~10, pp. 1222--1233, 1993.

\bibitem{yoshida2015ringshapefem}
E.~Yoshida, K.~Ayusawa, I.~G. Ramirez-Alpizar, K.~Harada, C.~Duriez, and
  A.~Kheddar, ``Simulation-based optimal motion planning for deformable
  object,'' in \emph{2015 IEEE Int. Workshop on Adv. Robotics and its Social
  Impacts (ARSO)}, 2015, pp. 1--6.

\bibitem{geometricallyexact2008}
A.~Theetten, L.~Grisoni, C.~Andriot, and B.~Barsky, ``Geometrically exact
  dynamic splines,'' \emph{Comput.-Aided Des.}, vol.~40, no.~1, pp. 35--48,
  2008.

\bibitem{umetani2014pbd}
N.~Umetani, R.~Schmidt, and J.~Stam, ``Position-based elastic rods,'' in
  \emph{Proc. of the ACM SIGGRAPH/Eurographics Symp. on Comput. Animation},
  ser. SCA '14.\hskip 1em plus 0.5em minus 0.4em\relax Eurographics Assoc.,
  2015, p. 21–30.

\bibitem{liu2023pbd}
F.~Liu, E.~Su, J.~Lu, M.~Li, and M.~C. Yip, ``Robotic manipulation of
  deformable rope-like objects using differentiable compliant position-based
  dynamics,'' \emph{IEEE Robotics Automation Lett.}, vol.~8, no.~7, pp.
  3964--3971, 2023.

\bibitem{berenson2013jacobian}
D.~Berenson, ``Manipulation of deformable objects without modeling and
  simulating deformation,'' in \emph{2013 IEEE/RSJ Int. Conf. on Intelligent
  Robotics and Systems}, 2013, pp. 4525--4532.

\bibitem{lv2020survey}
N.~Lv, J.~Liu, H.~Xia, J.~Ma, and X.~Yang, ``A review of techniques for
  modeling flexible cables,'' \emph{Comput.-Aided Des.}, vol. 122, p. 102826,
  2020.

\bibitem{kragic2021survey}
H.~Yin, A.~Varava, and D.~Kragic, ``Modeling, learning, perception, and control
  methods for deformable object manipulation,'' \emph{Sci. Robotics}, vol.~6,
  no.~54, p. eabd8803, 2021.

\bibitem{koessler2021icra}
A.~Koessler, N.~R. Filella, B.~Bouzgarrou, L.~Lequièvre, and J.-A.~C. Ramon,
  ``An efficient approach to closed-loop shape control of deformable objects
  using finite element models,'' in \emph{2021 IEEE Int. Conf. on Robotics and
  Automation (ICRA)}, 2021, pp. 1637--1643.

\bibitem{servin2008rigidbodysequence}
M.~Servin and C.~Lacoursière, ``Rigid body cable for virtual environments,''
  \emph{IEEE Trans. on Visualization and Comput. Graphics}, vol.~14, no.~4, pp.
  783--796, 2008.

\bibitem{lv2017massspring}
N.~Lv, J.~Liu, X.~Ding, J.~Liu, H.~Lin, and J.~Ma, ``Physically based real-time
  interactive assembly simulation of cable harness,'' \emph{J. of Manuf.
  Systems}, vol.~43, pp. 385--399, 2017, special Issue on the 12th Int. Conf.
  on Front. of Des. and Manuf.

\bibitem{moll2006minimalenergy}
M.~Moll and L.~Kavraki, ``Path planning for deformable linear objects,''
  \emph{IEEE Trans. on Robotics}, vol.~22, no.~4, pp. 625--636, 2006.

\bibitem{javdani2011}
S.~Javdani, S.~Tandon, J.~Tang, J.~F. O'Brien, and P.~Abbeel, ``Modeling and
  perception of deformable one-dimensional objects,'' in \emph{2011 IEEE Int.
  Conf. on Robotics and Automation}, 2011, pp. 1607--1614.

\bibitem{tamar2019rss}
A.~Wang, T.~Kurutach, K.~Liu, P.~Abbeel, and A.~Tamar, ``Learning robotic
  manipulation through visual planning and acting,'' in \emph{Proc. of
  Robotics: Sci. and Systems}, FreiburgimBreisgau, Germany, June 2019.

\bibitem{fiducial}
Y.~{Lai}, J.~{Poon}, G.~{Paul}, H.~{Han}, and T.~{Matsubara}, ``Probabilistic
  pose estimation of deformable linear objects,'' in \emph{2018 IEEE 14th Int.
  Conf. on Aut. Sci. and Eng. (CASE)}, 2018, pp. 471--476.

\bibitem{rubinstein2004cem}
R.~Y. Rubinstein and D.~P. Kroese, \emph{The Cross Entropy Method: A Unified
  Approach To Combinatorial Optimization, Monte-Carlo Simulation (Information
  Science and Statistics)}.\hskip 1em plus 0.5em minus 0.4em\relax Berlin,
  Heidelberg: Springer-Verlag, 2004.

\end{thebibliography}

\end{document}